%% file: main.tex
\documentclass[letterpaper]{article} %
\usepackage[submission]{aaai2026}  %
\usepackage{times}  %
\usepackage{helvet}  %
\usepackage{courier}  %
\usepackage[hyphens]{url}  %
\usepackage{graphicx} %
\usepackage{natbib}  %
\usepackage{caption} %
\usepackage{algorithm}
\usepackage{algorithmic}

\usepackage{newfloat}
\usepackage{listings}
\DeclareCaptionStyle{ruled}{labelfont=normalfont,labelsep=colon,strut=off} %
\floatstyle{ruled}
\newfloat{listing}{tb}{lst}{}
\floatname{listing}{Listing}
\def\showauthors@on{T} 

\title{Emergent Crowds Dynamics from Language-Driven Multi-Agent Interactions}

\author {
    Yibo Liu\textsuperscript{\rm 1}\equalcontrib,
    Liam Shatzel\textsuperscript{\rm 1}\equalcontrib,
    Brandon Haworth\textsuperscript{\rm 1}
    Teseo Schneider\textsuperscript{\rm 1}
}
\affiliations {
    \textsuperscript{\rm 1}University of Victoria
}

\input{99-maacros}

\begin{document}

\maketitle
\begin{figure}[!ht]
    \centering
    \includegraphics[width=0.9\linewidth]{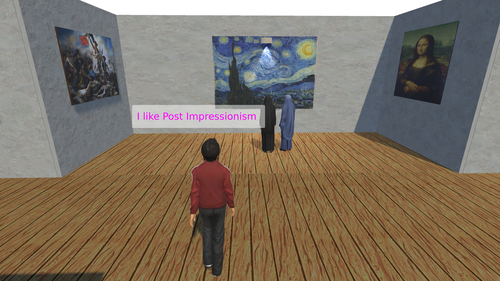}
    \includegraphics[width=0.9\linewidth]{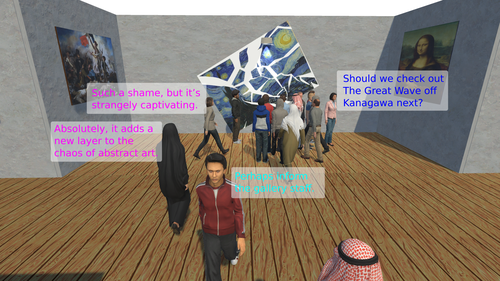}
    \caption{An animated scene illustrating public responses to a painting accident at a museum.}
    \label{fig:teaser}
\end{figure}

\begin{abstract}
\input{00-abstract}
\end{abstract}

\input{01-intro}

\input{02-related}

\input{03-method}

\input{04-results}

\input{05-conclusions}

\bibliography{ref}

\end{document}

%% file: 99-maacros.tex
\usepackage{xcolor}

\definecolor{TeseoColor}{rgb}{0.15, 0.68, 0.38}

\definecolor{Violet}{rgb}{.6, .35, .7}
\definecolor{Pink}{rgb}{1, 0, 1}
\definecolor{Yellow}{rgb}{1, 0.65, 0}

\definecolor{codegreen}{rgb}{0,0.6,0}
\definecolor{codegray}{rgb}{0.5,0.5,0.5}
\definecolor{codepurple}{rgb}{0.58,0,0.82}
\definecolor{backcolour}{rgb}{0.95,0.95,0.92}

\lstdefinestyle{mystyle}{
    backgroundcolor=\color{backcolour},   
    commentstyle=\color{codegreen},
    keywordstyle=\color{magenta},
    numberstyle=\tiny\color{codegray},
    stringstyle=\color{codepurple},
    basicstyle=\ttfamily\tiny,
    breakatwhitespace=false,         
    breaklines=true,                 
    captionpos=b,                    
    keepspaces=true,                 
    numbers=none,                    
    numbersep=0pt,    
    xleftmargin=0pt,
    xrightmargin=0pt,
    showspaces=false,                
    showstringspaces=false,
    showtabs=false,                  
    tabsize=2,
    breakatwhitespace=true
}

%% file: 00-abstract.tex
Animating and simulating crowds using an agent-based approach is a well-established area where every agent in the crowd is individually controlled such that global human-like behaviour emerges. We observe that human navigation and movement in crowds are often influenced by complex social and environmental interactions, driven mainly by language and dialogue. However, most existing work does not consider these dimensions and leads to animations where agent-agent and agent-environment interactions are largely limited to steering and fixed higher-level goal extrapolation.

We propose a novel method that exploits large language models (LLMs) to control agents’ movement. Our method has two main components: a dialogue system and language-driven navigation. We periodically query agent-centric LLMs conditioned on character personalities, roles, desires, and relationships to control the generation of inter-agent dialogue when necessitated by the spatial and social relationships with neighbouring agents. We then use the conversation and each agent's personality, emotional state, vision, and physical state to control the navigation and steering of each agent. Our model thus enables agents to make motion decisions based on both their perceptual inputs and the ongoing dialogue.

We validate our method in two complex scenarios that exemplify the interplay between social interactions, steering, and crowding. In these scenarios, we observe that grouping and ungrouping of agents automatically occur. Additionally, our experiments show that our method serves as an information-passing mechanism within the crowd. As a result, our framework produces more realistic crowd simulations, with emergent group behaviours arising naturally from any environmental setting. 

%% file: 01-intro.tex
\section{Introduction}

Animating and simulating crowds is a central problem in architecture, animation and entertainment, and more generally in human behaviour and scenario analysis. The main objective is to simulate and animate a crowd such that global human-like behaviour emerges. Many methods have emerged to achieve this that are largely differentiated in their approach, centralized vs. decentralized, agent-based vs. cellular automata vs. continuum, etc. Applications often dictate which approach is most useful given its pros and cons. Here we focus on directable, for animation purposes, decentralized agent-based simulation.

The general framework of a decentralized agent-based simulation follows three main phases: low frequency, deciding the global goal (e.g., the agent wants to leave the room); medium frequency, path planning with waypoints that the agent will \emph{loosely} follow to reach the goal (e.g., the path to reach the door and leave); high-frequency, steering where the agent is allowed to deviate from the path (e.g., to avoid collisions with other agents). Using these three stages, we can observe \emph{global emergent} phenomena; for instance, evacuation, where all agents must quickly leave an environment and navigate through bottlenecks. In the simulation, the single agent's behaviour influences the full crowd movement and leads to bottlenecks in the presence of small openings. 

We argue that, in normal circumstances (i.e., non-emergency situations), human movement is mostly driven by \emph{social} interaction with other humans, personal motivations, and local environmental/contextual affordances. For instance, while visiting your university clubs, you recognize your friend, join them, and jointly decide which stand to visit next (Section~\ref{sec:results}). This social interaction is usually not directly modelled and often leads to ``robot-like'' agent movements driven by orchestrated global goals and path planning for the agents involved. All agents move independently of each other and never naturally gather or re-plan their path based on the presence of other agents, memory, or natural discussion.

While large language models (LLMs) have been increasingly used in multi-agent settings to study social behaviour, most existing works focus solely on language and lack physical motion or interaction. We introduce our model to bridge this gap (Section~\ref{sec:method}). We propose a novel social multi-agent model that includes vision and language-based interaction between agents. Our model allows agents to collectively pick new goals, update the path planning, and change steering parameters to mimic human behaviour. We employ LLMs to control all locomotion phases, thus enabling generalizable crowd behaviours without scenario-specific modelling. Our model enables agents to make autonomous, decentralized decisions based on local visual observations and dialog, without relying on a centralized global prompt. By conditioning agent behaviour on individual agent personalites, emotional state, and agent-agent relationships (friends, family, enemy, etc.), as well as affording direct agent-agent communication channels, our framework represents a full agent simulation with a directable scenario and fine-tuned agent control. Agents can discuss, plan, see, avoid, direct, gather, and more, leading to complex social and contextual scenarios that previously would be impossible. We evaluate our method with two examples (Section~\ref{sec:results}); our experiments reveal that such micro-level language-motion interactions can give rise to emergent macro-level crowd behaviours.

%% file: 02-related.tex
\section{Related Work}
\paragraph{Crowds Simulation.}
Crowd simulation, the process of simulating and animating multiple interacting humans, has a long and vibrant literature with roots in cellular automata, agent-based models, flocking animation, and behavioural modelling~\cite{lemonari2022authoring,van2021algorithms,martinez2017modeling,badlerBook,DBLP:books/daglib/0030710}. A basic yet robust and extensible approach affords flocking and crowding behaviour generated by using simple rules~\cite{reynolds_flocks_1987}. The earliest explorations of this modelling and animation approach for humans incorporated basic collision prediction and grouping relationships~\cite{DBLP:books/daglib/0030710}. Physically-based methods emerged as a simple to implement, robust and effective approach for many use cases, dependent on the formulation of potential field functions mapping to internal agent motivations~\cite{helbing_social_1995}. More sophisticated models incorporated predictive collision avoidance~\cite{karamouzas2014universal,karamouzas_predictive_2009}. In parallel,
velocity space optimization formulations promise collision-free paths while producing emergent behaviours~\cite{van2011reciprocal,van2008reciprocal}.

\paragraph{Narrative Synthesis for Animation.}
Narrative synthesis has its roots in classic pre-generative text pre-LLM Natural Language Processing (NLP) and symbolic AI work. Early work related to agent animation synthesis built event libraries with even preconditions and author-set event goals to formulate event-driven narrative plans~\cite{shoulson2013event}. Methods that decouple user input and narrative map affordances to objects, allowing a planner to order the affordances such that they satisfy preconditions to form consistent narratives~\cite{kapadia2015computer}. This partial-order planner approach has been coupled with animation synthesis to generate scenes of agents carrying out roles and actions in a narrative arc consistent manner~\cite{kapadia2016canvas}. Converting the animation of narrative problem to that of resource allocation given some agents and environment affordance mappings, a resource allocation planner can synthesize day-long agent movements with steering for micro movement consistency~\cite{zhang2019multi}. This allows for the rapid dynamic population of a semantically labelled built-environment scenario.

\paragraph{Agent-based LLM-powered Simulation.}
Scalable LLM-driven multi-agent simulation is a likely new frontier of large-scale human behaviour simulation~\cite{adornetto2025generative,gao2024large}. While there is great promise in solving numerous classic problems (such as those faced by classic NLP-based narrative synthesis, or classic simplification issues in agent-based models (ABM) for behaviour simulation), LLMs may also present yet another behavioural model simplification now amplified by their black-box nature~\cite{larooij2025large}. Validation of these models increasingly complicates the already difficult problem of handling calibration, validation, and reproduction of ABMs.

Novel frameworks for LLM-driven agent-based simulation provide tools for city-scale scenario analysis where agents have memories, personalities, and diverse behaviours important for social investigations~\cite{bougie2025citysim}. Recent advances in LLM query scheduling and parallelization enable the simulation of several thousand agent-based models of individual macro behaviours in urban environments~\cite{yan2024opencity}. More detailed agent-agent and agent-environment simulation supports memory, reflection, and planning incorporated with basic shortest path construction for agent movement~\cite{park2023generative}. This allows agent locations to fundamentally and dynamically alter the environment as agent can now incorporate their new locale information (nearby agents, environment details, etc.) into their state.

The Parallel Information Aggregation via Neural Orchestration (PIANO) architecture combines micro and macro behavioural simulation in a Minecraft environment for civilization-scale modelling ($\sim 1000$ agents modelling social structures and information/religious exchange)~\cite{altera2024sid}. PIANO facilitates agent-agent and agent-human interaction in real-time at scale by running several modules or tasks concurrently and at varying timescales, respectively, to their natural use (slow long-term planning, immediate actions, regular memory retrieval, as-needed intent generation, etc).

\paragraph{LLMs for Crowds Simulation.}
Recent efforts have led to the incorporation of LLMs into the classic crowds synthesis problem, where macro behaviours (goals and intent) are incorporated with micro behaviours (steering and collision avoidance). This approach promises deep socio-cultural affordances combined with the ability to condition agents on personalities and scenarios with ease. One key issue in classic crowds is the parameterization of agents and the scenarios they operate in. This information can be learned in conditioned graphs that represent interactions in scenarios that allow prompt-based generation of new scenarios~\cite{ji2024text}. Recent work discretizes the action space of agents and learns the latent graph structure of agent-agent relations in scenarios. Using variational graph auto-encoders to learn agent-agent interaction as well as agent navigation action allows for the synthesis of new crowd scenarios. Conditioning this on a language model provides a natural way to synthesize new scenarios from text descriptions~\cite{panayiotou2025gen}. 
Our work provides a decentralized agent-based method for rich, unbounded agent-agent and agent-environment interactions with individual agent memory, detailed personality, emotion, and relationship conditioning that affords complex emergent phenomena expected in real crowds.

%% file: 03-method.tex
\section{Method}\label{sec:method}

With the observation that human movement is mostly driven by social/linguistic interaction with other humans, we propose a language-driven system that controls the behaviour and motion of each individual agent $a_i, i=1\dots,n$ in a crowd of $n$ agents (Section~\ref{sec:agent}).

Both our models (the language system and the language-driven movement) use an LLM for their decision-making. We provide them with the global state $S$ containing a description of the environment and the list of locations and possible goals (names and positions). To ensure that the LLM produces a reasonable and parsable output, we specify the format (Supplemental~\ref{app:format}), its role (Supplemental~\ref{app:format}), its task (Supplemental~\ref{app:format}), and provide a simple set of instructions (Supplemental~\ref{app:instructions}). We force the LLM to describe the image and analyze its output greatly improves its results, inspired by the chain-of-thought technique~\cite{10.5555/3600270.3602070}. If the LLM does not produce a valid output (i.e., it does not respect the format), we skip the frame. We note that this is an extremely rare event: less than 10 times in all our experiments.

\subsection{Agent}\label{sec:agent}

\begin{figure}
    \centering
    \includegraphics[trim=20 50 20 50, clip, width=0.6\linewidth]{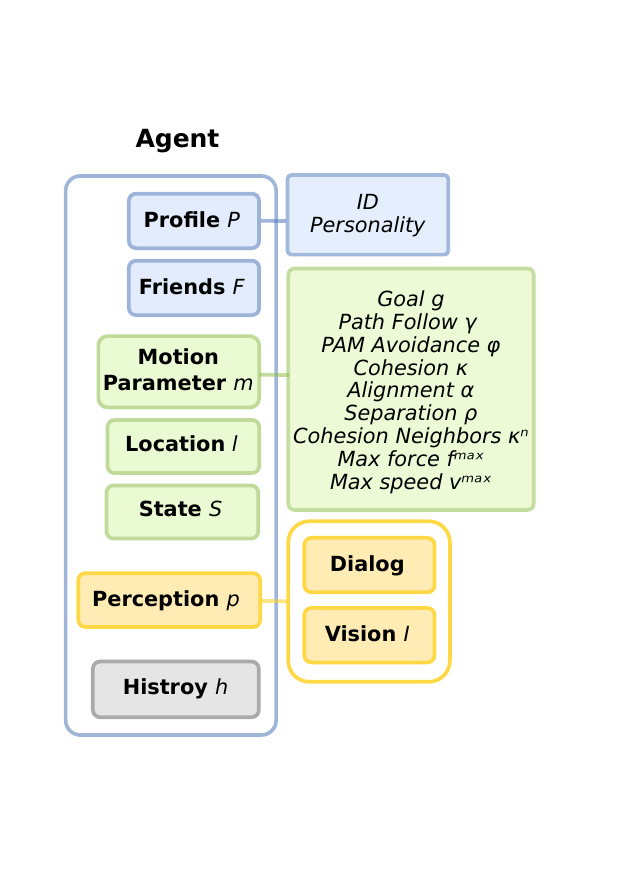}
    \caption{Overview of the agent's variables. In blue, we denote the static parts, in green the motion, and in yellow the perception.}
    \label{fig:agent}
\end{figure}

\begin{figure*}
    \centering
    \includegraphics[width=\linewidth]{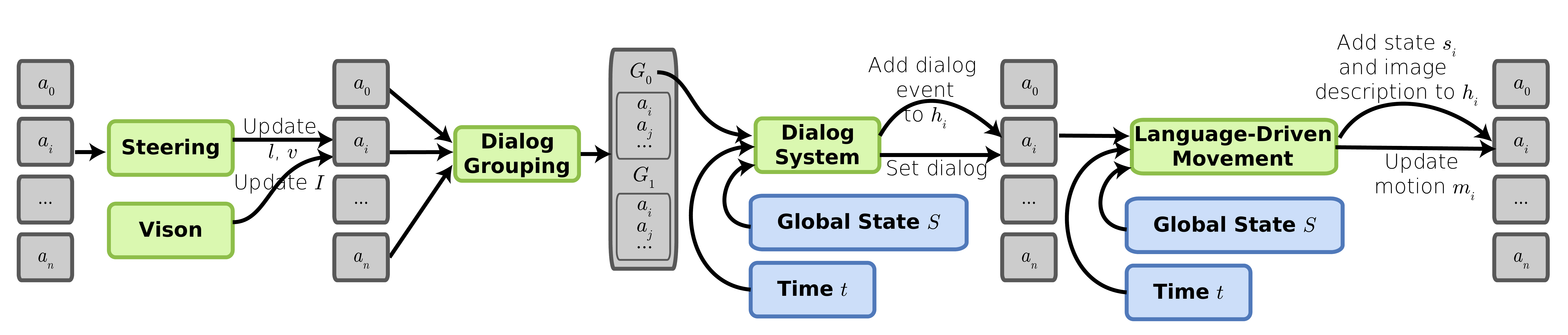}
    \caption{Overview of how our system updates the different parts of an agent. We first steer it, then update its perception, and finally update its motion, history, and state.}
    \label{fig:overview}
\end{figure*}

Every agent $a_i$ (Figure~\ref{fig:agent}) has a fixed profile $P_i$ containing its ID and personality (initialized with the MBTI type of personality \cite{myers1962mbti}), a list $F_i=\{f_i^j, j=1\dots,m\}$ of $m$ friends (note that friendship is not necessarily symmetric), a state $s_i$ (empty in the beginning), a location $\ell_i$, its perception of the world $p_i$, movement parameters $m_i$, and its history $h_i$ storing the past $k=50$ events (pair of time-stamp and four possible updated actions: position, dialog, vision, and movement) chronologically sorted. When the memory is full, we pop the oldest memory from the list.

The perception $p_i$  is composed of an image $I_i$ representing $a_i$ vision, and a dialog (Section~\ref{sec:dialog}). The movement parameters $m_i$ are composed of the current agent goal $g_i$ (randomly initialized) and steering parameters $d_i$. Our language-driven movement uses the agent's state and is responsible for updating it after every decision. We use the steering parameters $d_i=\{\gamma_i,\varphi_i,\kappa_i, \alpha_i, \kappa^n_i, \rho_i, f_i^{\max}, v_i^{\max}\}$ with weights: $\gamma_i$ for path-following, $\varphi_i$ for PAM avoidance, $\kappa_i$ for cohesion, $\alpha_i$ for alignment, and $\rho_i$ for separation. We also use a list of cohesion neighbours $\kappa^n_i$, and control the maximum force $f_i^{\max}$ and speed $v_i^{\max}$.
Since both the goals and steering parameters are controlled by our language-driven movement (Section~\ref{sec:ldn}), the whole motion path is influenced by the agent's state, its vision, and conversation.

\subsection{Overview}

For every agent $a_i$ and at every $t$ time steps in our animation (Figure~\ref{fig:overview}), we first update $a_i$ location $\ell_i$ and velocity $v_i$ using a force-based steering algorithm (Section~\ref{sec:steering}). We then update the agent perception $p_i$ by rendering the image $I_i$ from $a_i$ perspective and update their current dialog by grouping the agents (Section~\ref{sec:trigger}) and use our dialog system (Section~\ref{sec:dialog}) to generate a conversation. With the updated perception, we use our language-driven movement (Section~\ref{sec:ldn}) to obtain the new movement parameters $m_i$, new state $s_i$, and update its history $h_i$ by appending the summary of the decision process. Note that since both the dialog system and the language-driven movement use an LLM, which is slow, we call them every $k$ frames (we use $k=100$ in our experiments).

\subsection{Steering}\label{sec:steering}
To control the motion, we use the standard $A^\star$ algorithm and the string pulling algorithm to generate a series of waypoints~\cite{kallmann2016geometric}. In the steering, we combine boid~\cite{reynolds_flocks_1987} and PAM~\cite{karamouzas_predictive_2009}, which leads to a predictive force-based steering algorithm. Formally, for a timestep $t$ and an agent $a_i$ with location $\ell_i^t$ and velocity $v_i^t$, we define the
following force 
\[
f^{\gamma}_i=W_\gamma\gamma_i ((\ell_i^t-w)/\|\ell_i^t-w\| - v_i^h), 
\]
with $w$ the current waypoint and $W_\gamma=1$,
PAM force
cohesion force
\[
f^{\kappa}_i=\kappa_i\sum_{a_j\in\kappa^n_i} (\ell_i^t-\ell_j^t),
\]
alignment force
\[
f^{\alpha}_i=\alpha_i \sum_{a_j\in\kappa^n_i} v_j^t,
\]
and separation force
\[
f^{\rho}_i=\rho^i\sum_{a_j\in N_i}(\ell_i^t-\ell_j^t)e^{\rho_r/(\ell_i^t-\ell_j^t)},
\]
with $\rho_r=0.4$ and $N_i=\{a_j \mid \|\ell_i^t-\ell_j^t\|\leq \rho_r\}$. 
For every obstacle, we compute its bounding box radius $\rho_o$ and center $\ell_o$, and define an obstacle force
\[
f^o_i = W_o \sum_{o\in O}(\ell_i^t-\ell_o)/\|\ell_i^t-\ell_o\|,
\]
where $O$ is the list of obstacles at distance less than $\rho_o$ to $\ell_i^t$ and $W_o=10$. We then compute the total force
\[
F_i = \min(f_i^{\max}, f^{\varphi}_i+f^{\kappa}_i+f^{\alpha}_i+f^{\rho}_i+f^o_i),
\]
use an explicit time integrator to compute the new velocity
\[
v_i^{t+1}=\min(v_i^{\max}, v_i^{t}+\Delta t F_i),
\]
and update the location (which we append to the agent's history)
\[
\ell_i^{t+1}=\ell_i^{t}+\Delta t v_i^{t+1},
\]
where $\Delta t$ is the timestep.

\subsection{Dialog Grouping}\label{sec:trigger}
Since not all agents might participate in the same conversation, we collect all agents that are within 1.5 meters of each other. To avoid large, chaotic conversations, we limit the maximum number of agents in the conversation by a user-controlled parameter $g_{\max}$. In our experiments, we found that a maximum group size of 6 covers all scenarios. This produces a list of groups $G_i$ of agents that could participate in a conversation. Note that not all groups contain the same number of agents, and the number of groups varies during the animation.

\subsection{Dialog System}\label{sec:dialog}

The dialog system takes a single group $G_i$ of $m$ agents that could participate in a conversation and decides who in the group wants to join the conversation, whether to generate a conversation, and the content of the conversation. We exploit an LLM to generate natural dialog between different agents. We input to our dialog system (Figure~\ref{fig:overview}, middle) the current time $t$, global state $S$, and every agent $a_k\in G_i$ in the group. It outputs the dialog~\ref{conv:no-grp} between the agents in the group as an ordered list of pairs (agent id and dialog, see for instance~\ref{conv:grp}), which we then use to update the perception of every agent involved in the conversation. Note that we also append the current dialog (and time) to the event's history $h_i$.

\subsection{Language-Driven Movement}\label{sec:ldn}

We build a language-driven movement planner (Figure~\ref{fig:overview}, left) that takes the agent $a_i$, the current time $t$, and the global state $S$. With this input, it decides whether to change the agent's goal $g_i$, update the motion parameters $m_i$, or continue in the current state. 
Since the decision might affect the agent's state, as it produces a new state $s'_i$, it also summarizes the decision process, which we add to the agent's history $h_i$. Finally, the LLM also generates a short description of the image $I_i,$ which we also append to $h_i$.

%% file: 04-results.tex
\section{Experiments}\label{sec:results}

We test our method on two different scenarios where we show how agent movement is affected by the interaction and dialogues between the agents. Our scenes contain multiple agents, while in the second, we focus on interesting inter-agent interactions; we refer to the supplemental material for the video, animation, conversation and decision logs. Experiments were run on a workstation with an NVIDIA 3090 GPU, Intel I9 CPU, on Ubuntu OS.

\subsection{Scene 1: University Club Fair}
In this setup, we simulate a university club fair, a social event where conversations naturally occur. The scene includes 50 agents, each initialized with a randomly assigned MBTI personality type from the 16 standard categories. Their initial goals and starting positions are also randomized, with all characters entering from one side of the fair. We separate agents from movable (30 agents) and club staff that do not move (20 agents). Additionally, at the international center, the agents can get free T-shirts, and \emph{only} a few club staff (7 agents) are aware of it.

We run the simulation runs at 50 frames per second (FPS), and use our method (Section~\ref{sec:method}) every 100 frames to generate dialogue and update the agent's movement parameters.

\begin{figure*}
    \centering\footnotesize
    \includegraphics[width=.24\linewidth]{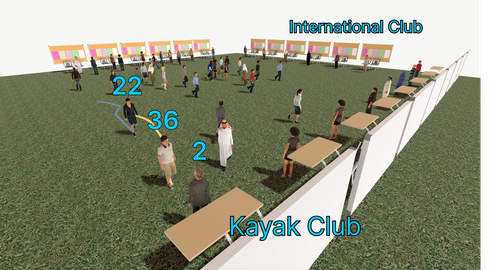}\hfill
    \includegraphics[width=.24\linewidth]{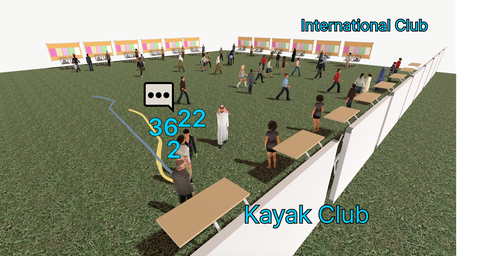}\hfill
    \includegraphics[width=.24\linewidth]{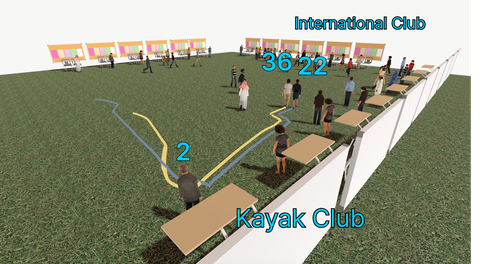}\hfill
    \includegraphics[width=.24\linewidth]{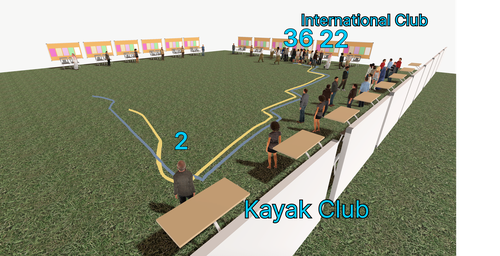}\par
    \parbox{.24\linewidth}{\centering Frame 199}\hfill
    \parbox{.24\linewidth}{\centering Frame 384}\hfill
    \parbox{.24\linewidth}{\centering Frame 654}\hfill
    \parbox{.24\linewidth}{\centering Frame 1294}
    \caption{Example of how two characters meet, chat with an agent, take a collective decision, and walk to a new goal \emph{together}.}
    \label{fig:grouping}
\end{figure*}

\paragraph{Grouping through Implicit Motion Parameters.}
Figure \ref{fig:grouping} illustrates the emergence of grouping behaviour. At frame 199 (first figure), $a_{36}$ is en route to the kayaking club, where $a_{22}$ is already present. By frame 384, $a_{36}$ arrives and engages in a conversation with the club staff $a_{2}$ and $a_{22}$ (second figure). After Conversation~\ref{conv:grp}, where $a_{2}$ informs them about free T-shirts at the international center, they begin walking side by side (third figure), and by frame 1294, they arrive at the international center (fourth figure). Their trajectory shows that they travel together the entire way.

\begin{lstlisting}[language=Python, caption={From the suggestion of $a_{2}$, $a_{22}$ and $a_{36}$ decide to visit the  international center to get free T-shirts.}, label={conv:grp}]
2:  "Welcome to the kayaking club! Interested in joining?"
36: "Yes, what's available here?"
2:  "We organize weekly kayaking trips. It's fun and social."
36: "Sounds great! Anything else happening today?"
2:  "The international center is giving out free T-shirts."
36: "Good to know, I'll visit after. Thanks!"
22: "Let's check it out together after this."
\end{lstlisting}

Following the conversation, our system made specific adjustments to the motion parameters for both agents $a_{22}$ and $a_{36}$. Both of their goals changed from \textit{kayak club} to \textit{international center}.
Within $a_{22}$ steering parameters, cohesion $\kappa_{22}$ increased from 0.0 to 1.0, and $a_{36}$ was added to its cohesion neighbours list $\kappa^n_{22}$. Similarly for $a_{36}$, $\kappa_{36}$ increase from 0.0 to 0.8, the alignment $\alpha_{22}$ increased from 0.0 to 0.7 (as $a_{36}$ follows $a_{22}$), and $a_{22}$ was added to $\kappa^n_{36}$.

\begin{figure}
    \centering\footnotesize
    \includegraphics[width=.48\linewidth]{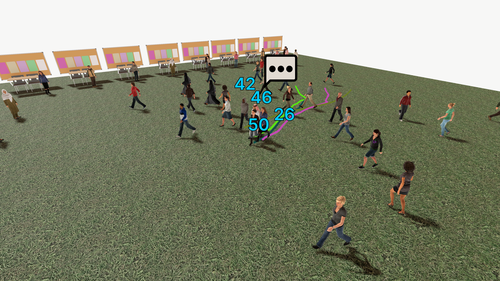}\hfill
    \includegraphics[width=.48\linewidth]{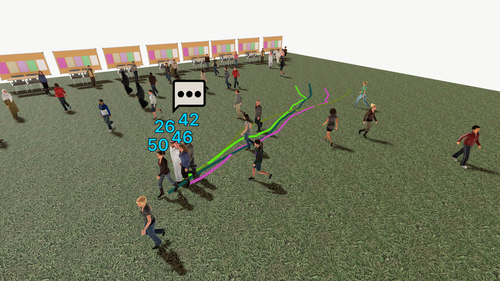}
    \parbox{.48\linewidth}{\centering Frame 304}\hfill
    \parbox{.48\linewidth}{\centering Frame 429}
    \caption{Our system supports the natural formation and dynamic evolution of larger agent groups. An agent asked to join a spontaneously formed group of three.}
    \label{fig:large-grp}
\end{figure}

In the same scene, we observe (Conversation~\ref{conv:join}) $a_{42}$ asking a group of three agents to join (Figure~\ref{fig:large-grp}, left) and the subsequent formation of a group consisting of four agents (Figure~\ref{fig:large-grp}, right). This example demonstrates how conversation can lead agents to synchronize their goals and walking behaviour. It also indicates that agents are capable of interpreting intent and updating motion parameters in a hierarchical and context-aware manner, showing evidence of their understanding of motion dynamics. 

\begin{lstlisting}[language=Python, caption={$a_{42}$ joining the group of $a_{26}$, $a_{46}$, and $a_{50}$.}, label={conv:join}]
26: "Hey, we're almost at the gardening club. What do you all want to check out first?"
50: "I'm curious about the plant care tips they offer."
46: "I've heard they have unique species on display."
42: "Mind if I join you? I was heading to the international center, but this sounds fun."
26: "Of course, the more the merrier!"
42: "Great, let's see if they have workshops too."
50: "Absolutely, workshops would be interesting. Let's find out."
\end{lstlisting}

Leveraging LLMs as parameter controllers, our model regulates movement at a low level while implicitly capturing high-level behaviors like pairing and grouping, without explicitly modeling collective dynamics.

\begin{figure}
    \centering\footnotesize
    \includegraphics[width=.48\linewidth]{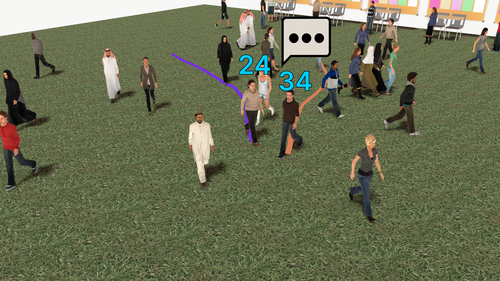}\hfill
    \includegraphics[width=.48\linewidth]{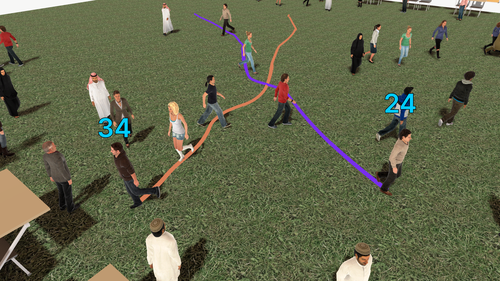}
    \parbox{.48\linewidth}{\centering Frame 329}\hfill
    \parbox{.48\linewidth}{\centering Frame 559}
    \caption{Two friends meet, but ultimately they decide not to walk together.}
    \label{fig:flexgrouping}
\end{figure}

\paragraph{Soft Grouping.}
Not all conversations result in grouping (Conversation~\ref{conv:no-grp}). As shown in Figure~\ref{fig:flexgrouping}, we observe a case where two friends ($a_{24}$ and $a_{34}$) meet while walking, have a brief chat, but ultimately choose not to walk together toward the same goal. This outcome is also strongly influenced by the agents' personalities, as reflected in the reasoning provided by the LLM (Conversation~\ref{conv:no-grp-exp}).

\begin{lstlisting}[language=Python, caption={Because of their personalities, $a_{24}$ and $a_{34}$, decide to not walk together.}, label={conv:no-grp}]
24: "Hey 34, heading to the music club. Want to join?"
34: "Hi 24, I'm actually on my way to the caving club now."
24: "Music club's nearby! We can check it out quickly."
34: "Tempting, but I promised to meet someone at caving club."
24: "Alright, let's catch up later then. Text me!"
34: "Will do. Have fun at the music club!"
\end{lstlisting}

\begin{lstlisting}[breakindent=0pt, caption={LLM reasoning for Conversation~\ref{conv:no-grp}.}, label={conv:no-grp-exp}]
In this scenario, Agent 24 and Agent 34 are close friends located near each other at a university club fair. Agent 24, an ENFJ, feels excited about being near the music club and having a friend (Agent 34) nearby, which can amplify their outgoing social behaviour. Agent 34, an ISTJ, is generally more reserved and focused on reaching the caving club, their current goal.

Given their close proximity and established friendship, it's highly probable they'd have a brief conversation. Social dynamics suggest Agent 24 might try to persuade Agent 34 to visit the music club together, leveraging their excitement and the social environment of the fair. However, because Agent 34 is an ISTJ and goal-oriented, they might not be easily swayed from their current objective. Instead, they might agree to meet up later after visiting their respective clubs. Since both agents are currently walking and clearly visible to each other, they'd likely pause briefly for this interaction. This conversation will not change their immediate goals, but it reinforces their friendship and sets the stage for potential future activities together.
\end{lstlisting}

This shows that agents consider personality traits, enabling natural, flexible grouping, and reason from both psychological and physical perspectives.

\begin{figure}
    \centering\footnotesize
    \includegraphics[width=.48\linewidth]{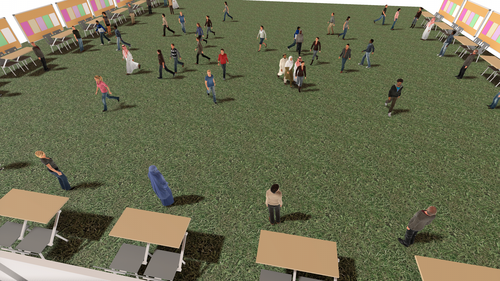}\hfill
    \includegraphics[width=.48\linewidth]{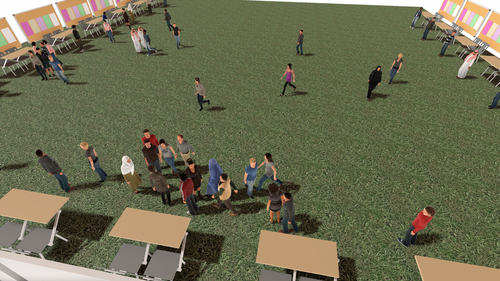}
    \parbox{.48\linewidth}{\centering Frame 354}\hfill
    \parbox{.48\linewidth}{\centering Frame 799}
    \caption{Because of the free T-shirts available at the international center, the initial agents' distribution (left) naturally converges into a crowd (right) by the end of the animation.}
    \label{fig:gathering}
\end{figure}

\paragraph{Emergent Gathering via Conversational Coupling.}
Figure~\ref{fig:gathering} illustrates how conversational information passing leads to emergent gathering behavior. At frame 354, agents are evenly distributed across the space, while by frame 799, a noticeable gathering has formed at the international center. The information about free T-shirt distribution is initially known only to a few staff agents. Without explicitly scripting this knowledge for all agents, it propagates naturally through dialogue (e.g., Conversation~\ref{conv:grp}). Each agent individually adjusts its motion based on LLM reasoning, and conversational interactions socially couple their internal states. This implicit coordination enables synchronized, group-level movement to emerge from local, language-driven interactions.

\subsection{Scene 2: Accident at Museum}
In this setup, we simulate an accident occurring in a museum, where the painting ``The Starry Night'' falls from the wall and becomes cracked. This event is introduced in the environment prompt at frame 799. The scene allows us to observe how agents respond to both a dynamic environment and a specific, unexpected event.

The simulation includes 30 agents, each initialized with a randomly assigned a random individual art preference. Their initial goals and starting positions are also randomized, with all agents entering through the museum’s main entrance.

\begin{figure}
    \centering\footnotesize
    \includegraphics[width=.48\linewidth]{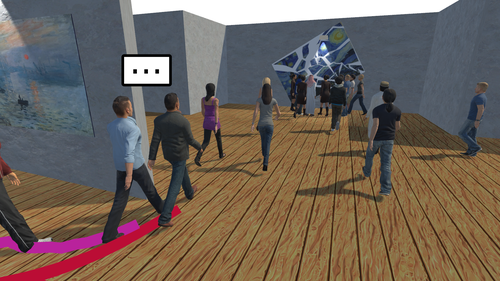}\hfill
    \includegraphics[width=.48\linewidth]{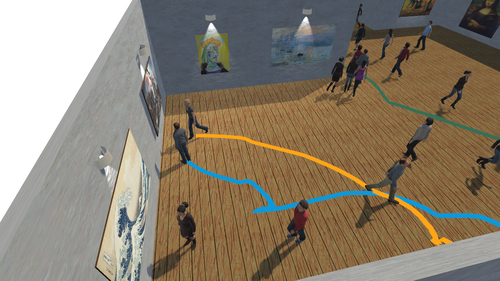}
    \parbox{.48\linewidth}{\centering Frame 1199}\hfill
    \parbox{.48\linewidth}{\centering Frame 1139}    \caption{Example of various reactions to the museum painting falling accident. Left, the agents have conversations while walking to the accident area. Right, the agents ignore the accident chaos.}
    \label{fig:museum1}
\end{figure}

\paragraph{Various Unscripted Reactions.}
This experiment reveals that our method can lead to a range of diverse, unscripted reactions to the accident event; many agents walk toward the damaged painting, forming a visible flow (Figure~\ref{fig:museum1} left). 

Their behaviour can be explained by their dialog (Conversation~\ref{conv:curious}) and the movement planner.
\begin{lstlisting}[language=Python, caption={$a_{24}$ and $a_{28}$, decide to walk towards the damaged painting.}, label={conv:curious}]
24: "Did you hear that crash? 'The Starry Night' fell!"
28: "Yes, I did. It's concerning. We should check it out together."
24: "Absolutely. I hope it isn't damaged badly."
28: "Me too. Such a masterpiece deserves better protection."
24: "It's a shame if it suffered. Let's hurry and see."
28: "Agreed. After that, maybe explore more surreal pieces?"
24: "Sounds perfect. Let's go!"
\end{lstlisting}

In the lower image, $a_{2}$, $a_{20}$, $a_{21}$ remain focused on their current artworks and are not drawn to the chaos caused by the fallen painting, as visible by Conversation~\ref{conv:ignore}. 

\begin{lstlisting}[breakindent=0pt, caption={LLM reasoning for $a_{20}$.}, label={conv:ignore}]
Having just arrived at 'Liberty Leading the People,' the painting's bold colors and evocative scene capture your attention, aligning with your preference for art that tells a story. Despite the earlier disturbance, there is no visible chaos or urgent situation in the gallery area, suggesting the incident has settled. With friends clustered nearby and given the absence of immediate distress signals, there is no pressing need to leave or change plans radically. However, curiosity and enthusiasm for the gallery's offerings remain high, prompting the possibility of exploring another vibrant piece. For now, you will maintain calm, allowing time to absorb the painting, before potentially moving to another artwork if friends decide to explore further.
\end{lstlisting}

\begin{figure}
    \centering\footnotesize
    \includegraphics[width=.48\linewidth]{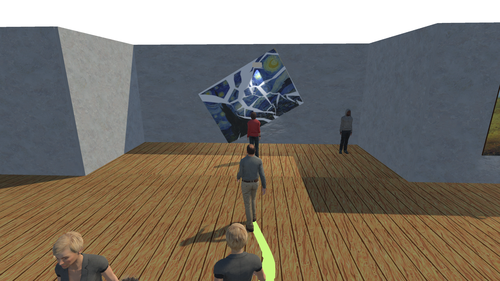}\hfill
    \includegraphics[width=.48\linewidth]{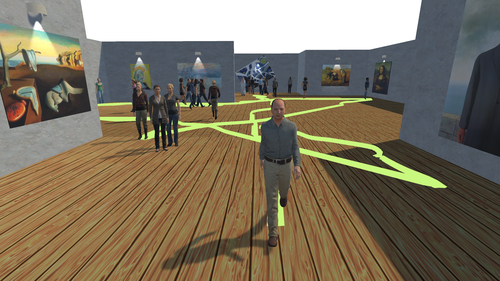}
    \parbox{.48\linewidth}{\centering Frame 5164}\hfill
    \parbox{.48\linewidth}{\centering Frame 5809}
    \caption{$a_{2}$ looks at the accident and decides to report to staff (left) and navigate to the entrance to find a staff member (right). }
    \label{fig:museum2}
\end{figure}

$a_{2}$ decides to report the damage to the staff at the museum entrance after observing the fallen painting (Conversation~\ref{conv:broken}) at frame 5164 (Figure~\ref{fig:museum2} left); then navigates towards the entrance (\ref{fig:museum2} right); explained by Conversation~\ref{conv:broken-why}. This event leads to an updated state of $a_{2}$ to \texttt{Increased anxiety due to witnessing the art accident} which affects its decision-making and can change in response to visual information.

\begin{lstlisting}[breakindent=0pt, caption={Vison description from $a_{2}$.}, label={conv:broken}]
In front of me is 'The Starry Night,' now damaged and shattered on the floor. The vivid brushstrokes lie in disarray, a tragic contrast to their usual swirling serenity. The scene is startling, with friend ID 4 standing approximately 2 meters ahead at 0 degrees, observing the damage. The fragmented colors create a chaotic visual.
\end{lstlisting}

\begin{lstlisting}[breakindent=0pt, caption={Reasoning analysis of movement decision for $a_{2}$.}, label={conv:broken-why}]
The sight of 'The Starry Night' shattered on the floor enhances my mild anxiety. As someone who finds deep stories within brushstrokes, the destruction of this iconic artwork is unsettling. This emotional response, coupled with the presence of friend ID 4, prompts a decision to change the destination. I aim to discuss this incident with others and perhaps inform the gallery staff, as lingering here won't alter the painting's state. Thus, I'll adjust my destination towards the 'Entrance_Exit' to potentially report the issue. Maintaining my current walking style allows me to focus on navigating through the gallery without delay, given the urgency of the situation.
\end{lstlisting}

These examples show our model produces diverse, context-aware behaviours in response to environmental changes, enabling dynamic animation and scriptless narrative continuation.

\begin{figure}
    \centering\footnotesize
    \includegraphics[width=.48\linewidth]{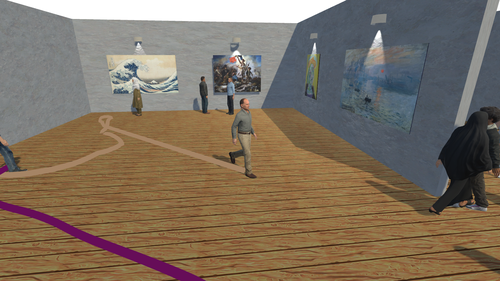}\hfill
    \includegraphics[width=.48\linewidth]{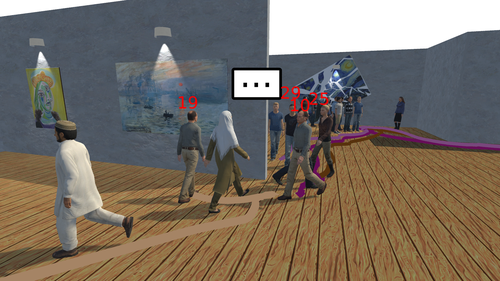}
    \parbox{.48\linewidth}{\centering Frame 1444}\hfill
    \parbox{.48\linewidth}{\centering Frame 1909}
    \includegraphics[width=.48\linewidth]{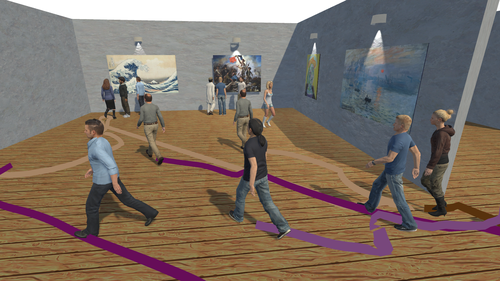}\hfill
    \includegraphics[width=.48\linewidth]{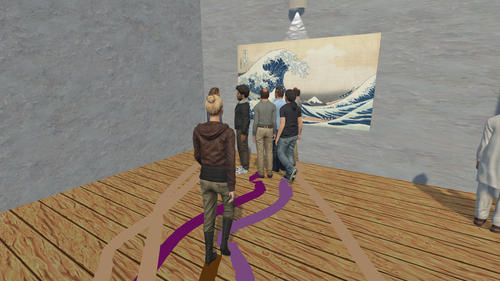}
    \parbox{.48\linewidth}{\centering Frame 2079}\hfill
    \parbox{.48\linewidth}{\centering Frame 2349}
    \caption{Example of conversation changing agent's decision for movement.}
    \label{fig:museum3}
\end{figure}

\paragraph{Language Influence on Movement Decisions}
\figurename~\ref{fig:museum3} illustrates how $a_{19}$ changes his destination as a result of a conversation with other agents. At frame 1444, $a_{19}$ is walking toward the painting ``The Starry Night''. At frame 1909, he encounters $a_{10}$, $a_{25}$, and $a_{29}$. They tell a bad joke (Conversation~\ref{conv:joke}) about the broken painting, which causes $a_{19}$ to change his mind—deciding not to visit ``{The Starry Night}'', but instead to join the others in heading toward ``{The Great Wave off Kanagawa}''. As a result, we see that by frame 2079, they begin walking in that direction, and by frame 2349, they are nearly there.

\begin{lstlisting}[language=Python, caption={Bad joke that change $a_{19}$'s mind at Frame 1994.}, label={conv:joke}]
10: "Did you hear that noise? 'The Starry Night' just fell!"
19: "I saw it too! Such a shame, but it's strangely captivating."
25: "Absolutely, it adds a new layer to the chaos of abstract art."
29: "Should we check out 'The Great Wave off Kanagawa' next?"
10: "Yes, let s go. It's nearby and fits our style."
19: "Agreed, and maybe we can discuss this incident along the way."
25: "Perfect, let's explore more vibrant artworks together!"
\end{lstlisting}

Integrating conversation into a multi-agent animation system shapes movement and fosters behavioral diversity.

\begin{figure}
    \centering\footnotesize
    \includegraphics[width=.48\linewidth]{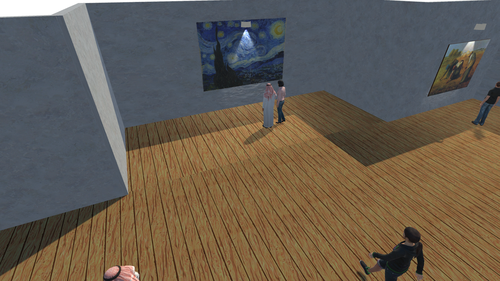}\hfill
    \includegraphics[width=.48\linewidth]{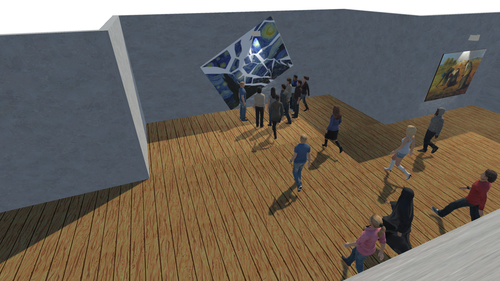}
    \parbox{.48\linewidth}{\centering Frame 724}\hfill
    \parbox{.48\linewidth}{\centering Frame 1199}
    \includegraphics[width=.48\linewidth]{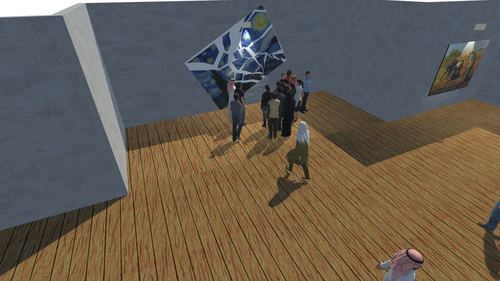}\hfill
    \includegraphics[width=.48\linewidth]{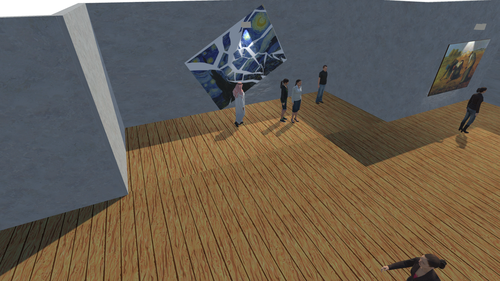}
    \parbox{.48\linewidth}{\centering Frame 1484}\hfill
    \parbox{.48\linewidth}{\centering Frame 2789}
    \caption{Emergent gathering and dispersal behaviour, before falling frame 724 and after frames 1199, 1584 and 2789.}
    \label{fig:museum4}
\end{figure}

\paragraph{Emergence of Gathering and Dispersal.}
In \figurename~\ref{fig:museum4}, we can observe high-level crowd behaviours such as gathering and dispersal around the fallen painting. At frame 724, before the accident occurs, agents are naturally distributed throughout the museum, while some engage with different artworks, and others explore. At frame 1199 we can see agenets walking towards the damaged painting. By frame 1484, several agents have gathered around the damaged painting to investigate what happened. At frame 2519, the gathering gradually dissolves.

These dynamics emerge naturally from agent interactions, demonstrating our model’s ability to produce environment-sensitive behaviours without explicit scripting.

%% file: 05-conclusions.tex
\section{Conclusion and Future Work}

We introduced a novel LLM-guided navigation system for multi-agent animation. Our model enables character interactions to emerge in a realistic and unscripted manner, allowing for richer and more natural behaviours, such as gathering, coordination, and message propagation through physical proximity. 

Our approach avoids scripting; each agent uses the LLM for localized decision-making, allowing complex group behaviours to emerge from decentralized interactions. Tested in two scenarios, our system demonstrates how LLMs can drive diverse, context-aware behaviours through motion-level control, enabling emergent dynamics in spatially grounded environments.

Our method uses standard navigation and steering, allowing it to be a drop-in replacement in existing systems. This feature also comes with all the limitations of existing methods. For instance, the navigation and steering might be conflicting, or the agent struggles to stop moving when close (but not too close) to a goal. In the future, we plan to investigate if using an LLM to model agents might enable a more holistic agent's control and possibly sidestep some of the problems.

Finally, as we use a distributed LLM API, our system is slow (around 5 seconds per query). It would be interesting to investigate to usage of local (smaller) models, where the queries can be much faster.